\documentclass{bmvc2k}

\title{Learning to Project for Cross-Task Knowledge Distillation} 
\addauthor{Dylan Auty$^\ast$}{dylan.auty12@imperial.ac.uk}{1}
\addauthor{Roy Miles$^\ast$}{r.miles18@imperial.ac.uk}{1}
\addauthor{Benedikt Kolbeinsson$^\ast$}{benedikt.kolbeinsson15@imperial.ac.uk}{1}
\addauthor{Krystian Mikolajczyk}{k.mikolajczyk@imperial.ac.uk}{1}

\addinstitution{
MatchLab \\
Imperial College London \\
South Kensington, UK \\
}

\newcommand\blfootnote[1]{%
  \begingroup
  \renewcommand\thefootnote{}\footnote{#1}%
  \addtocounter{footnote}{-1}%
  \endgroup
}

\runninghead{Auty et al.}{Learning to Project for Cross-Task Knowledge Distillation}

\usepackage{amsmath}
\usepackage{amssymb}
\usepackage{booktabs}
\usepackage{caption}
\usepackage{subcaption}
\usepackage{svg}
\usepackage[utf8]{inputenc} %
\usepackage[T1]{fontenc}    %
\usepackage{url}            %
\usepackage{amsfonts}       %
\usepackage{nicefrac}       %
\usepackage{microtype}      %

\usepackage{multirow}
\usepackage{epsfig}
\usepackage{colortbl}
\usepackage{stackengine}
\usepackage{xcolor}
\usepackage[most]{tcolorbox}
\usepackage{refcount}
\usepackage{bm}

\usepackage{tikz}
\usepackage{tabularx}

\newcommand*{\belowrulesepcolor}[1]{%
  \noalign{%
    \kern-\belowrulesep 
    \begingroup 
      \color{#1}%
      \hrule height\belowrulesep 
    \endgroup 
  }%
} 
\newcommand*{\aboverulesepcolor}[1]{%
  \noalign{%
    \begingroup 
      \color{#1}%
      \hrule height\aboverulesep 
    \endgroup 
    \kern-\aboverulesep 
  }%
} 

\definecolor{mygrey}{RGB}{230,230,230} 
\definecolor{myblue}{RGB}{230,230,255}
\definecolor{myred}{RGB}{251,231,231}
\definecolor{myyellow}{RGB}{255,255,175}
\definecolor{ourcolor}{RGB}{191,223,191}

\definecolor{d2d}{RGB}{207,225,28}
\definecolor{s2d}{RGB}{50,181,121}
\definecolor{c2d}{RGB}{49,100,141}
\definecolor{r2d}{RGB}{68,2,86}

\definecolor{blue_eqn}{RGB}{93,124,230}
\definecolor{red_eqn}{RGB}{217,88,71}

\newcommand{\lpnormv}[1]{\lVert #1 \rVert}

\def \alambicdeit {\includegraphics[width=0.02\linewidth]{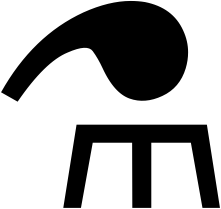}\xspace}

\newcolumntype{Y}{>{\centering\arraybackslash}X}

\begin{document}

\maketitle

\blfootnote{$^\ast$ The authors contributed equally to this paper}
\begin{abstract}

    Traditional knowledge distillation (KD) relies on a proficient teacher trained on the target task, which is not always available.
    In this setting, cross-task distillation can be used, enabling the use of any teacher model trained on a different task. However, many KD methods prove ineffective when applied to this cross-task setting. To address this limitation, we propose a simple modification: the use of an inverted projection. We show that this drop-in replacement for a standard projector is effective by learning to disregard any task-specific features which might degrade the student's performance.
    We find that this simple modification is sufficient for extending many KD methods to the cross-task setting, where the teacher and student tasks can be very different.
    In doing so, we obtain up to a 1.9\% improvement in the cross-task setting compared to the traditional projection, at no additional cost.
    Our method can obtain significant performance improvements (up to 7\%) when using even a randomly-initialised teacher on various tasks such as depth estimation, image translation, and semantic segmentation, despite the lack of any learned knowledge to transfer.
    To provide conceptual and analytical insights into this result, we show that using an inverted projection allows the distillation loss to be decomposed into a knowledge transfer and a spectral regularisation component. 
    Through this analysis we are additionally able to propose a novel regularisation loss that allows teacher-free distillation, enabling performance improvements of up to 2.3\% on ImageNet with no additional training costs.
\end{abstract}


\begin{figure}
\centering
\subfigure[Traditional same-task knowledge distillation that projects to the teacher feature space.]{\label{fig:overview-same-task}\includegraphics[width=0.45\textwidth]{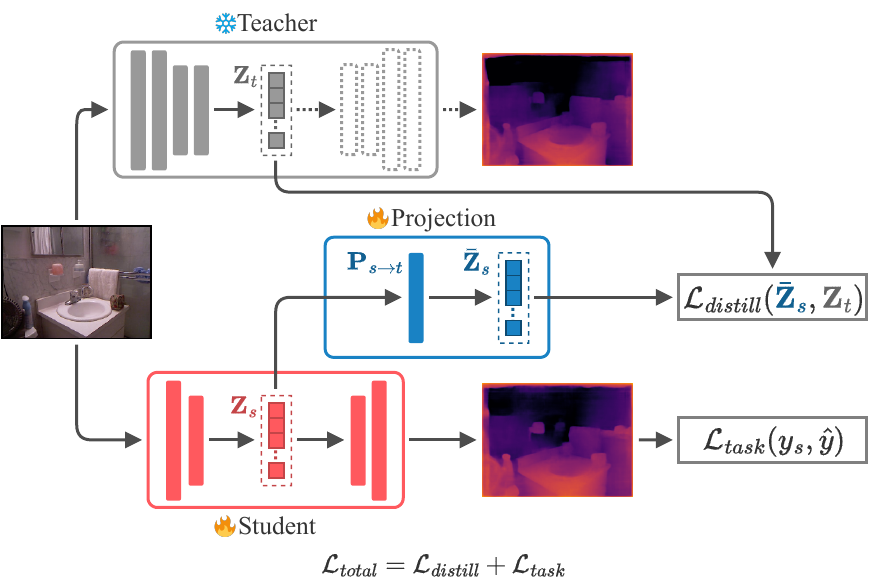}}
\hfill
\subfigure[Cross-task knowledge distillation using our inverted projection.]{\label{fig:overview-ours-cross-task}\includegraphics[width=0.45\textwidth]{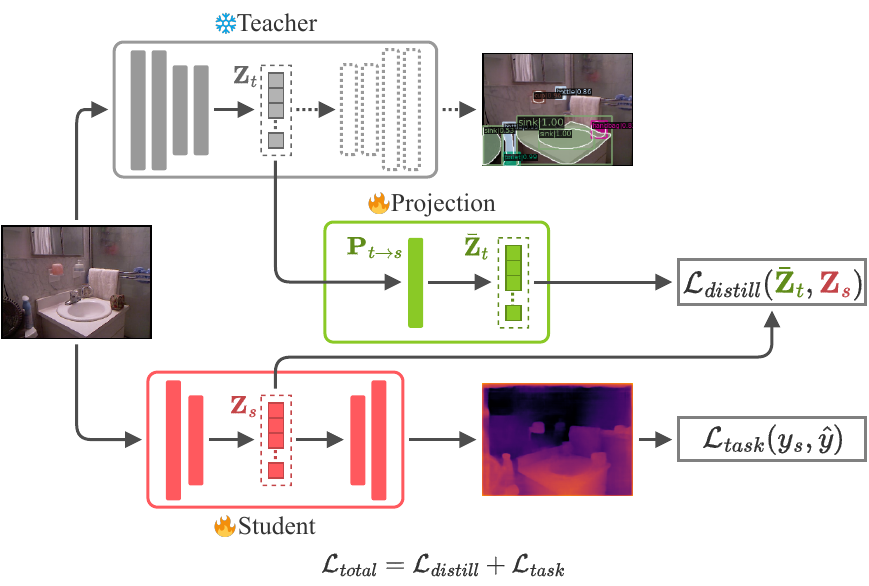}}
\vspace{1em}
\caption{\textbf{Our cross-task knowledge distillation pipeline}, where a student model is trained on a target task with the aid of a frozen teacher that is pretrained on a \textbf{different} task.
    Compared to standard same-task feature distillation (fig. \ref{fig:overview-same-task}), our cross-task approach uses an \textit{inverted projector} (fig. \ref{fig:overview-ours-cross-task}) which is able to discard irrelevant task-specific features from the different-task teacher.
    The loss comprises a feature distillation loss $\mathcal{L}_{distill}$ that matches the student features with the projected teacher features, and a task-specific supervised loss $\mathcal{L}_{task}$ applied only to the student model's output for the target task.}
    \vspace{-1em}
    \label{fig:overview}
\end{figure}

\section{Introduction}
\label{sec:introduction}

Knowledge distillation (KD) has emerged as a very effective tool for training small and efficient models~\cite{Tian2019ContrastiveDistillation, Miles2023MobileVOS:Distillation, Fang2021ContrastiveDistillationv2, bhardwaj2019dream, miles2020cascaded, lopezpaz2016unifying}. 
It leverages the pre-trained knowledge of a much larger (teacher) model to guide and enhance the training process of a significantly smaller (student) model. 
Since its inception, KD has been applied to a wide variety of tasks in the computer vision~\cite{Chen2021DistillingReview}, audio~\cite{chen2021distillingaudio}
and language~\cite{DistilBert2019} domains, enabling the deployment of models across many embedded devices.

However, existing approaches for KD are often limited to the cases where the teacher shares the same task with the student~\cite{huang2022knowledge, chen2022dearkd, Hinton2015DistillingNetwork, Tian2019ContrastiveDistillation, Chen2020WassersteinDistillation, Chen2021DistillingReview}. 
This is very restrictive since there are many applications where there is simply no suitable pretrained teacher available due to, for example, the lack of any large annotated training data. This problem commonly arises for tasks that require expensive human annotation~\cite{mccormac_scenenet_2017}, such as in robotics~\cite{james2019simtoreal}, or where the collection of data is prohibitive for other reasons, such as in the medical~\cite{ronneberger2015unet, Komorowski_2023_CVPR} and aerial domains~\cite{wang2020tartanair,fonder2019mid,kolbeinsson2023ddos}. 
In these cases, it is not possible to train a suitable teacher for the target task, therefore we propose a \textbf{cross-task knowledge distillation}. In the cross-task KD setup, a teacher model trained for a \textbf{different} task can be used to improve the student performance. 
This setting is well-suited for tasks lacking a task-specific pretrained teacher, as it allows for any other off-the-shelf pretrained model to be used to improve the student model performance instead.
It is also increasingly relevant as the compute and data costs to train large models increases.
Similarly, data labelling may be cheaper for one task than another, e.g. training a model using a cheaply-labelled auxiliary task is very common in active learning~\cite{baldridge_active, cost_acc_active, ijcai2017p0261} and federated learning~\cite{ahn2022federated}.

We show that the traditional methods for same-task KD fail in this new and more general cross-task setting since they transfer domain-specific knowledge, which is associated with the \textit{teacher's} task.
Therefore, while they increase the student's performance in the traditional same-task setting, they degrade it in the cross-task scenario. We propose the use of an inverted projection to address this problem. We find that this modification is very effective in the cross-task setting due to its suppression of task-specific information. Most notably, we can obtain up to a 7.47\% performance improvement by distilling from a teacher trained on various different tasks.
We demonstrate that this simple drop-in replacement enables many KD methods to adapt to the cross-task setting, and we show consistent improvements across various tasks including depth estimation, segmentation, and image translation.

To obtain more insights into the underlying mechanism of the inverted projector, we explore the training dynamics of its weights.
We find that the least-significant singular vectors of the teacher's features are suppressed in cases where there is a significant task gap, which indicates that these singular vectors tend to be more task-specific.
Based on this observation we show that the suppression of singular vectors by the projector naturally leads to a decoupling of the distillation loss into a knowledge transfer and spectral regularisation component. 
This enables us to derive a \textbf{cheap spectral regularisation loss.} We describe this loss as a \textit{teacher-free distillation} method since it explicitly exploits the emergent regularisation component from cross-task distillation.
The new loss makes it possible to efficiently achieve performance competitive with many state-of-the-art KD methods without the need for any pre-trained teachers, with a 3.2\% relative improvement over the baseline model on ImageNet-1K.
In summary, our contributions are given as follows:

\begin{itemize}
    \item We propose a simple modification to standard KD that enables cross-task distillation: a learnable inverted projection.
    \item We show consistent and substantial performance improvements in the cross-task setting of up to 7\% through extensive experiments.
    \item By analysing the training dynamics of the projector weights, we are able to decouple a knowledge transfer and spectral regularisation component. We use this to derive a teacher-free regularisation loss that obtains up to 8\% improvement over the baseline with no additional training cost.
\end{itemize}

\section{Related Work}
\label{sec:related_work}

\textbf{Knowledge distillation (KD)} is a technique that involves transferring the knowledge from a large teacher model to a smaller student model, aiming to improve the performance of the student on the target task. It has become increasingly popular in recent years for the deployment of models on resource constrained devices, such as mobile phones, and has been applied in image classification~\cite{Hinton2015DistillingNetwork, Chen2020WassersteinDistillation}, semantic segmentation~\cite{Liu2019StructuredSegmentation}, video object segmentation~\cite{Miles2023MobileVOS:Distillation}, and natural language processing~\cite{Sanh2019DistilBERTLighter}. Existing literature has extensively explored various distillation pipelines~\cite{Malinin2020EnsembleDistillation, Attention2020PayingDistillation, Ren2022Co-advise:Distillation} along with both empirical and theoretically motivated loss formulations~\cite{Hinton2015DistillingNetwork, Zhao2022DecoupledDistillation, Chen2020WassersteinDistillation, Miles2022InformationDistillation} that can facilitate the knowledge transfer process. However, these conventional methods still predominantly focus on same-task distillation~\cite{Tian2019ContrastiveDistillation, Chen2021DistillingReview}, wherein the student and teacher models are trained on the same task. 
There are many applications where there are no pre-trained off-the-shelf teacher models available, thus motivating the need to perform \textit{cross-task distillation}. Some prior works have pursued cross-task distillation both as a generalisation of the knowledge transfer occurring in traditional knowledge distillation \cite{Ye2020DistillingMatching} and because of the observation that some tasks will naturally tend to share information \cite{yuanCKDCrossTaskKnowledge2020,Yang2022Cross-TaskRecommendation}.
CrossDistil~\cite{Yang2022Cross-TaskRecommendation} was one of the first to partially explore this new setting by introducing a quadruplet loss, calibration term, and an error correction mechanism, however knowledge was distilled between the task-specific decoder heads of a multi-task model with shared encoder weights, rather than between two fully-separate models. 
ProC-KD~\cite{Li2022Prototype-guidedModels} transfer local-level object knowledge for large-scale cross-task distillation, while \cite{Ye2020DistillingMatching} construct a relational embedding for the loss.
\cite{yuanCKDCrossTaskKnowledge2020} perform cross-task KD to augment text-to-image generation using image semantic encoders, but the proposed method is tightly coupled with the model architecture at each stage of the pipeline.
In contrast to these works, we propose a very simple extension to the typical feature distillation pipeline that enables the distillation of knowledge cross-task across a wide range of settings.

\noindent\textbf{Transfer learning and domain adaptation} are widely studied areas in machine learning that leverage the knowledge acquired by a pretrained model on one task to enhance the performance on a different, yet related task~\cite{zhuang2020comprehensive}. This paradigm has demonstrated significant success in various fields, especially in computer vision~\cite{zamir_taskonomy_2018,evci2022head2toe} and natural language processing~\cite{raffel2020exploring,sung2022vl}, by reducing the training time and data requirements in the target domain.
Most existing transfer learning or domain adaptation algorithms attempt to align the feature representations across the two domains. This can be achieved by minimising some statistical discrepancy between the two spaces~\cite{Greenfeld2019RobustCriterion} or introducing additional adversarial losses~\cite{Deng2021AdversarialRepresentations}. More recent works~\cite{pmlrchen19i} have shown a spectral divide between the domain-specific and domain-agnostic features. This is where the large singular values of the features can generalise across domains, whereas the small singular values are domain-specific. This observation has led to follow-up works~\cite{pmlrchen19i, batch_shrinkage_nips19, raab_bridging_2020} by proposing spectral alignment and normalisation techniques. We take a similar approach for transferring knowledge between different tasks, but in the context of knowledge distillation, where an additional capacity gap between the source and target task models exists. This work also enables a more concrete bridge between the field of transfer learning and knowledge distillation.

\noindent\textbf{Multi-task learning.} There are many cases where jointly training on multiple tasks or modalities can improve not only the generality of models~\cite{radford2021learning} but also the single-task performance. 
For example, monocular depth estimation has been shown to share knowledge with other tasks, such as semantic segmentation \cite{ramamonjisoa_sharpnet:_2019, jiao_look_2018, bai_monocular_2019, xing_roiformer_2022, wang_sdc-depth_2020,auty_monocular_2022,kolbeinsson2024ucorr}. Intuitively, this follows for other task pairs; for instance, both semantic segmentation and classification target the semantics within an image.
Unfortunately, multitask models are often too large and expensive to run on resource-constrained devices~\cite{kirillov2023segment}. Additionally, jointly learning multiple tasks with a small model can degrade the downstream performance, as additional tasks or objectives can conflict with the target task when there is insufficient capacity in the student to optimise for both~\cite{fifty2021efficiently}.

\vspace{-1.5em}
\section{Method}
\label{sec:method}


\subsection{Cross-task Feature Distillation}
Cross-task distillation \label{sec:subsec:method:cross-task-feature-distillation}
is motivated by the intuitive and demonstrated overlap in useful information between different tasks (see section \ref{sec:related_work}).
We use feature-space distillation, which aims to align the feature spaces of a student model and a teacher model. To do this, a learnable projection is used to map the features from one model into the feature space of the other.
However, in the cross-task case, where the teacher model has been trained for a significantly different task to the student model's target task, there are specific issues to contend with that traditional KD methods do not address.
We introduce a novel inverted projection that is well-suited to the cross-task setting, in contrast to the traditional projection~\cite{romero_fitnets_2015, Tian2019ContrastiveDistillation} which is better suited to the same-task setting. 

\subsection{Importance of Feature Projection}%
\label{sec:subsec:why-does-proj-space-matter}
In the traditional same-task setting, the teacher model is already pre-trained for the student's target task, so it is desirable for the student to match features as closely as possible with those produced by the teacher.
In this case, the task-specific knowledge is helpful in improving the student performance.
However, for the cross-task setting, the teacher is trained on a different task to the student.
As detailed in section \ref{sec:related_work}, there is likely at least some shared knowledge between different tasks.
The issue in the cross-task knowledge distillation setting is how to extract only the \textit{task-agnostic} knowledge and the knowledge \textit{shared} between tasks, all while ignoring the irrelevant features produced by teacher. This last point is especially important for smaller student models as they do not have the capacity to effectively learn the union of two very different feature spaces.
A projection layer is often used in knowledge distillation to match the student's feature dimensions with those of the teacher~\cite{romero_fitnets_2015, Tian2019ContrastiveDistillation}. Although recent works have highlighted the importance of the projector in improving the efficacy of same-task distillation~\cite{Chen2022ImprovedEnsemble, miles2024understanding, orthogonalKD2024}, they have proven ineffective when there is cross-task information present. We propose a modification of the projection in which we instead map from the teacher space onto the student space. We describe this as \textit{inverting} the projector, and we find that it enables the suppression of irrelevant (task-specific) features.
We show that this inverted projector can effectively discard these irrelevant features if needed. If this were to be used in the traditional same-task setting, the discarding of these features would be detrimental, but in the cross-task setting, it is actively desirable.

\begin{table}[t]
\centering
\resizebox{\textwidth}{!}{%
\begin{tabular}{ll|lll|lll|lll|lll}
\toprule
\belowrulesepcolor{mygrey}
        \rowcolor{mygrey} \multicolumn{2}{r|}{\textbf{Teacher task} $\longrightarrow$} & \multicolumn{3}{c|}{{\color{d2d} $\blacksquare$ } \textbf{Depth}}  & 
         \multicolumn{3}{c|}{{\color{s2d} $\blacksquare$ } \textbf{Instance Seg.}}  & 
         \multicolumn{3}{c|}{{\color{c2d} $\blacksquare$ } \textbf{Classification}}  &
         \multicolumn{3}{c}{{\color{r2d} $\blacksquare$ } \textbf{Random}}  \\
\rowcolor{mygrey} KD Method &
  Projection type &
  $\delta_1 \uparrow$ &
  Abs. $\downarrow$ &
  RMS $\downarrow$ &
  $\delta_1 \uparrow$ &
  Abs. $\downarrow$ &
  RMS $\downarrow$ &
  $\delta_1 \uparrow$ &
  Abs. $\downarrow$ &
  RMS $\downarrow$ &
  $\delta_1 \uparrow$ &
  Abs. $\downarrow$ &
  RMS $\downarrow$ \\ 
\aboverulesepcolor{mygrey}
\midrule
\multicolumn{2}{c|}{\textit{No teacher (baseline)}} &
  \stackengine{2pt}{\textit{0.845}}{$_{\pm 0.007}$}{U}{c}{F}{T}{S} &
  \stackengine{2pt}{\textit{0.127}}{$_{\pm 0.003}$}{U}{c}{F}{T}{S} &
  \stackengine{2pt}{\textit{0.440}}{$_{\pm 0.005}$}{U}{c}{F}{T}{S} &
  \stackengine{2pt}{\textit{0.845}}{$_{\pm 0.007}$}{U}{c}{F}{T}{S} &
  \stackengine{2pt}{\textit{0.127}}{$_{\pm 0.003}$}{U}{c}{F}{T}{S} &
  \stackengine{2pt}{\textit{0.440}}{$_{\pm 0.005}$}{U}{c}{F}{T}{S} &
  \stackengine{2pt}{\textit{0.845}}{$_{\pm 0.007}$}{U}{c}{F}{T}{S} &
  \stackengine{2pt}{\textit{0.127}}{$_{\pm 0.003}$}{U}{c}{F}{T}{S} &
  \stackengine{2pt}{\textit{0.440}}{$_{\pm 0.005}$}{U}{c}{F}{T}{S} &
  \stackengine{2pt}{\textit{0.845}}{$_{\pm 0.007}$}{U}{c}{F}{T}{S} &
  \stackengine{2pt}{\textit{0.127}}{$_{\pm 0.003}$}{U}{c}{F}{T}{S} &
  \stackengine{2pt}{\textit{0.440}}{$_{\pm 0.005}$}{U}{c}{F}{T}{S} \\
\midrule
\multirow{3}{1.95cm}{FitNets~\cite{romero_fitnets_2015} \\
    \textit{\footnotesize ICLR 2015}} &
  Traditional &
  \textbf{0.868} &
  \textbf{0.117} &
  \textbf{0.406} &
  \textbf{0.855} &
  \textbf{0.122} &
  \textbf{0.425} &
  0.845 &
  0.125 &
  0.439 &
  0.828 &
  0.134 &
  0.455 \\
 &
  Inverted (ours) &
  0.849 &
  0.124 &
  0.432 &
  0.851 &
  0.124 &
  0.431 &
  \textbf{0.850} &
  \textbf{0.124} &
  \textbf{0.434} &
  \textbf{0.851} &
  \textbf{0.124} &
  \textbf{0.431} \\
 &
  \textit{Improvement} &
  \cellcolor[HTML]{FFB0B0}\textit{-2.17\%} &
  \cellcolor[HTML]{FF1717}\textit{-6.35\%} &
  \cellcolor[HTML]{FF1212}\textit{-6.49\%} &
  \cellcolor[HTML]{FFF0F0}\textit{-0.41\%} &
  \cellcolor[HTML]{FFBEBE}\textit{-1.78\%} &
  \cellcolor[HTML]{FFCFCF}\textit{-1.31\%} &
  \cellcolor[HTML]{EDFFED}\textit{0.50\%} &
  \cellcolor[HTML]{ECFFEC}\textit{0.53\%} &
  \cellcolor[HTML]{CFFFCF}\textit{1.34\%} &
  \cellcolor[HTML]{97FF97}\textit{2.86\%} &
  \cellcolor[HTML]{00FF00}\textit{7.47\%} &
  \cellcolor[HTML]{42FF42}\textit{5.20\%} \\
\midrule
\multirow{3}{1.95cm}{AT~\cite{Zagoruyko2019PayingTransfer} \\
    \textit{\footnotesize ICLR 2017}} &
  Traditional &
  \textbf{0.856} &
  \textbf{0.122} &
  0.426 &
  0.852 &
  0.123 &
  0.431 &
  0.850 &
  0.125 &
  0.433 &
  \textbf{0.857} &
  \textbf{0.121} &
  \textbf{0.428} \\
 &
  Inverted (ours) &
  \textbf{0.856} &
  \textbf{0.122} &
  \textbf{0.425} &
  \textbf{0.855} &
  \textbf{0.121} &
  \textbf{0.429} &
  \textbf{0.853} &
  \textbf{0.123} &
  \textbf{0.430} &
  \textbf{0.857} &
  0.122 &
  \textbf{0.428} \\
 &
  \textit{Improvement} &
  \cellcolor[HTML]{FFFBFB}\textit{-0.11\%} &
  \cellcolor[HTML]{FFFCFC}\textit{-0.08\%} &
  \cellcolor[HTML]{FFFFFF}\textit{0.02\%} &
  \cellcolor[HTML]{F0FFF0}\textit{0.42\%} &
  \cellcolor[HTML]{CDFFCD}\textit{1.38\%} &
  \cellcolor[HTML]{ECFFEC}\textit{0.53\%} &
  \cellcolor[HTML]{F3FFF3}\textit{0.35\%} &
  \cellcolor[HTML]{C5FFC5}\textit{1.61\%} &
  \cellcolor[HTML]{E3FFE3}\textit{0.79\%} &
  \cellcolor[HTML]{FEFFFE}\textit{0.05\%} &
  \cellcolor[HTML]{FFF0F0}\textit{-0.83\%} &
  \cellcolor[HTML]{FCFFFC}\textit{0.09\%} \\ \midrule
\multirow{3}{1.95cm}{PKT~\cite{Passalis2018LearningTransfer} \\
    \textit{\footnotesize ECCV 2018}} &
  Traditional &
  \textbf{0.854} &
  \textbf{0.122} &
  0.429 &
  \textbf{0.857} &
  \textbf{0.123} &
  \textbf{0.427} &
  0.851 &
  0.124 &
  0.432 &
  0.856 &
  0.123 &
  0.429 \\
 &
 Inverted (ours) &
  \textbf{0.854} &
  \textbf{0.122} &
  \textbf{0.427} &
  0.854 &
  \textbf{0.123} &
  0.429 &
  \textbf{0.853} &
  \textbf{0.123} &
  \textbf{0.431} &
  \textbf{0.858} &
  \textbf{0.122} &
  \textbf{0.426} \\
 &
  \textit{Improvement} &
  \cellcolor[HTML]{FEFFFE}\textit{0.04\%} &
  \cellcolor[HTML]{FFF9F9}\textit{-0.16\%} &
  \cellcolor[HTML]{F0FFF0}\textit{0.42\%} &
  \cellcolor[HTML]{FFF2F2}\textit{-0.34\%} &
  \cellcolor[HTML]{FFFCFC}\textit{-0.08\%} &
  \cellcolor[HTML]{FFEEEE}\textit{-0.44\%} &
  \cellcolor[HTML]{F7FFF7}\textit{0.25\%} &
  \cellcolor[HTML]{D1FFD1}\textit{1.29\%} &
  \cellcolor[HTML]{F5FFF5}\textit{0.30\%} &
  \cellcolor[HTML]{F5FFF5}\textit{0.29\%} &
  \cellcolor[HTML]{D3FFD3}\textit{1.22\%} &
  \cellcolor[HTML]{E1FFE1}\textit{0.84\%} \\
\midrule
\multirow{3}{1.95cm}{Ensemble~\cite{Chen2022ImprovedEnsemble} \\
    \textit{\footnotesize NeurIPS 2022}} &
  Traditional &
  \textbf{0.861} &
  \textbf{0.119} &
  \textbf{0.416} &
  \textbf{0.856} &
  \textbf{0.122} &
  \textbf{0.425} &
  \textbf{0.852} &
  \textbf{0.124} &
  \textbf{0.431} &
  0.835 &
  0.128 &
  0.446 \\
 &
  Inverted (ours) &
  0.849 &
  0.124 &
  0.433 &
  0.848 &
  0.124 &
  0.435 &
  0.847 &
  0.125 &
  0.437 &
  \textbf{0.849} &
  \textbf{0.124} &
  \textbf{0.432} \\
 &
  \textit{Improvement} &
  \cellcolor[HTML]{FFC9C9}\textit{-1.46\%} &
  \cellcolor[HTML]{FF5656}\textit{-4.64\%} &
  \cellcolor[HTML]{FF6969}\textit{-4.11\%} &
  \cellcolor[HTML]{FFDCDC}\textit{-0.95\%} &
  \cellcolor[HTML]{FFC3C3}\textit{-1.64\%} &
  \cellcolor[HTML]{FFB0B0}\textit{-2.16\%} &
  \cellcolor[HTML]{FFE7E7}\textit{-0.63\%} &
  \cellcolor[HTML]{FFF0F0}\textit{-0.89\%} &
  \cellcolor[HTML]{FFCFCF}\textit{-1.30\%} &
  \cellcolor[HTML]{C0FFC0}\textit{1.74\%} &
  \cellcolor[HTML]{9BFF9B}\textit{2.75\%} &
  \cellcolor[HTML]{91FF91}\textit{3.03\%} \\ 
\bottomrule
\end{tabular}
}
\vspace{1em}
\caption{\textbf{Cross-task distillation to a depth estimation student model using similar ({\color{d2d} $\blacksquare$}) and dissimilar ({\color{r2d} $\blacksquare$}) teacher tasks},
showing the \textbf{increasing} effect of our inverted projection as similarity between teacher and student tasks \textbf{decreases}. 
We use our inverted projector with four different KD methods to show its general applicability.
The inverted projector outperforms traditional projections in the cross-task case for which it is designed, but always produces a performance improvement over the baseline (no distillation) regardless of the teacher task.
{\color{d2d} $\blacksquare$ }{\color{s2d} $\blacksquare$ }{\color{c2d} $\blacksquare$ }{\color{r2d} $\blacksquare$} colour map denotes \textit{decreasing} student-teacher task similarity.}
\label{tab:main_cross_task_distillation}
\end{table}

\subsection{Setup and Training loss}
\label{sec:subsec:setup}
Our cross-task knowledge distillation pipeline is shown in figure \ref{fig:overview-ours-cross-task}. 
It consists of a trainable student model, which is to be trained on a given target task, and a frozen teacher model, which is pre-trained on a different task.
This setup is in contrast to the traditional same-task knowledge distillation setting, which is shown in figure \ref{fig:overview-same-task}. Both the student and teacher models receive the same input image, and their respective encoders produce features $\mathbf{Z}_s$ and $\mathbf{Z}_t$ respectively. 
A learnable linear projection matrix $\mathbf{P}$ is used to project $\mathbf{Z}_t$ to the dimensions of $\mathbf{Z}_s$, giving $\bar{\mathbf{Z}}_t = \mathbf{Z}_t\mathbf{P}$. 
A distance function $d$ is then used between the student features and the projected teacher features:
\begin{equation} 
    \mathcal{L}_{distill} =
    d(\mathbf{Z}_s, \mathbf{Z}_t\mathbf{P}),
    \label{eq:distill-loss}
\end{equation}
where $d$ can be any distance metric, such as the L2 loss used by FitNets~\cite{romero_fitnets_2015} or the attention mapping described by AT~\cite{Attention2020PayingDistillation}.  
In addition to this loss, we also use a task-specific supervision loss between the student model's output $y_s$ and the ground truth labels $y$ to ensure the student's output aligns with the target task. 
Since the teacher's output is not used, we only perform a forward pass through its encoder in order to reduce the training compute required.
The final loss is given by:
\begin{equation}
    \mathcal{L}_{total} = \mathcal{L}_{task}(y_s, y) + \mathcal{L}_{distill}(\bar{\mathbf{Z}}_t, \mathbf{Z}_s),
    \label{eq:total_loss}
\end{equation}

where $\mathcal{L}_{task}$ is the downstream target-task loss. For example, for depth estimation it will be a pixel-wise loss with the ground truth depth, and for image classification it will be a cross entropy term.

\vspace{-0.5em}
\subsection{Decoupled Feature Distillation} 
\label{sec:subsec:decoupled}

\begin{figure}[!t]
    \centering
    \subfigure{\includegraphics[width=0.32\textwidth]{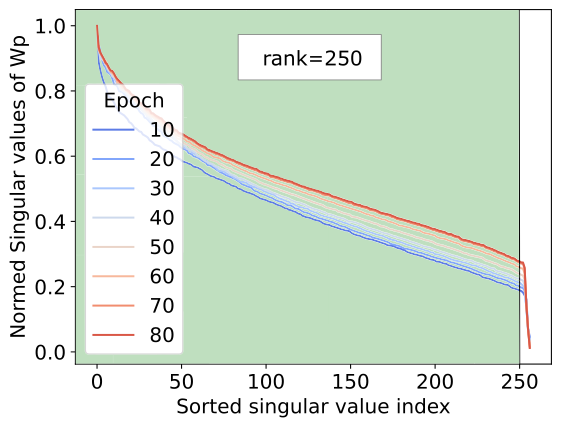}}
    \subfigure{\includegraphics[width=0.32\textwidth]{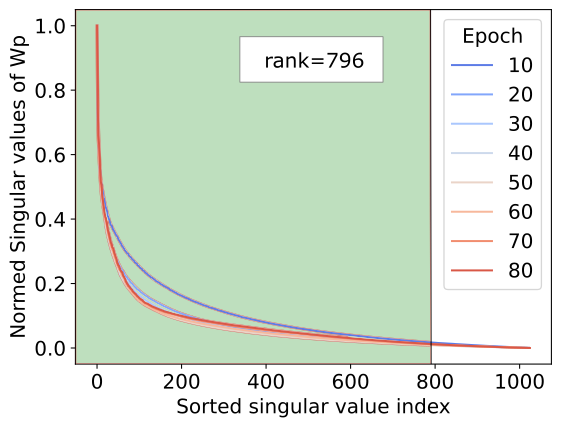}}
    \subfigure{\includegraphics[width=0.32\textwidth]{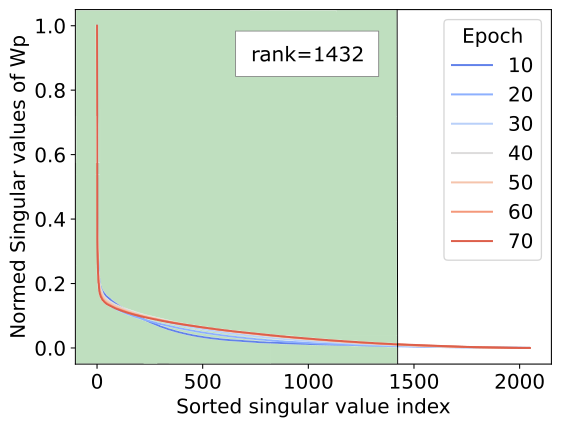}}
    \subfigure[{\color{d2d} $\blacksquare$ } Seg. $\rightarrow$ Seg.]{\includegraphics[width=0.32\textwidth]{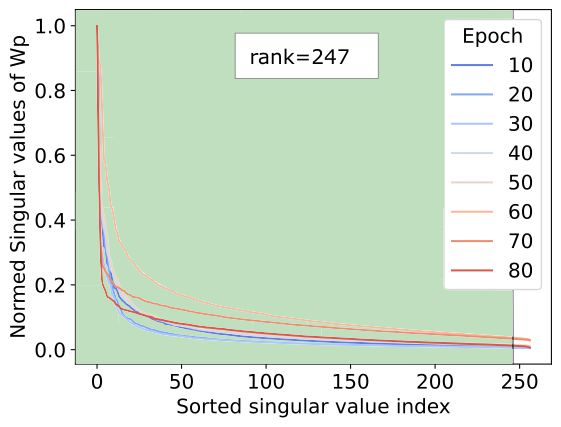}}
    \subfigure[{\color{c2d} $\blacksquare$ } Class. $\rightarrow$ Seg.]{\includegraphics[width=0.32\textwidth]{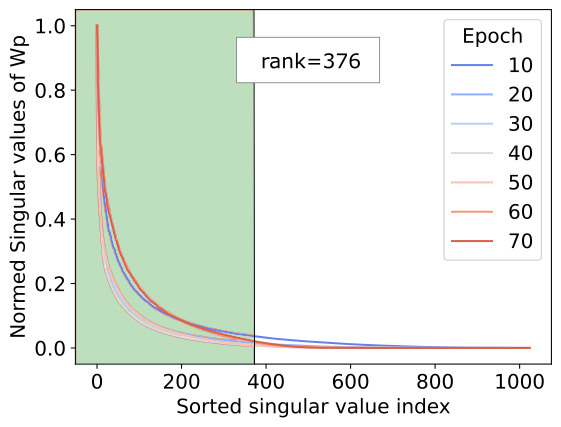}}
    \subfigure[{\color{r2d} $\blacksquare$ } Random $\rightarrow$ Seg.]{\includegraphics[width=0.32\textwidth]{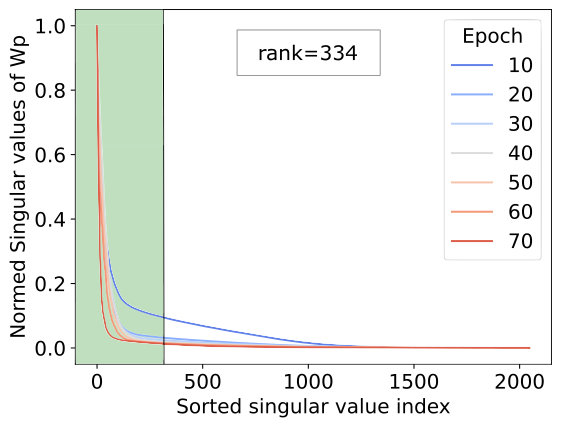}}
    \vspace{0.5em}
    \caption{\textbf{Evolution of singular values} of the projection matrix $\mathbf{P}$ under different cross-task settings and projector types. Green area highlights the rank of $\mathbf{P}$.
    The projection tends towards a higher rank either when using the traditional projection or when using the same or similar-task teacher. The low-rank when using our inverted projection in the cross-task setting allows irrelevant features to be filtered out, if necessary for the task pair.
    \textbf{Top row}: traditional projection, \textbf{Bottom row}: our inverted projection.}
    \label{fig:eigplots}
\end{figure}

To obtain analytical insights into the consequences of projecting the teacher features, we take the case where $\mathcal{L}_{distill}$ is a simple L2 loss between the student's features $\mathbf{Z}_s$ and the projected teacher features $\bar{\mathbf{Z}}_t$.
We perform singular value decomposition on both, i.e. $\bar{\mathbf{Z}}_t = \bar{\mathbf{U}}\bar{\pmb{\Sigma}} \bar{\mathbf{V}}^T$ and $\mathbf{Z}_s = \mathbf{U}\pmb{\Sigma} \mathbf{V}^T$. 
The cross-task setting requires that our inverted projection learns to discard the irrelevant task-specific features from the teacher model. 
This can be implemented using a low-rank projection of the features.
However, we observe that a low-rank projection naturally emerges in the cross-task setting when using our inverted projector. In fact, this emergence is even more prominent when there is a significant task gap (see section \ref{sec:experiments}).
Using this low-rank property, we can express $\bar{\mathbf{Z}}_t$ using a truncated SVD, i.e. keeping few non-zero singular values. Substituting this into our $\mathcal{L}_{distill}$ with an L2 loss, we can then decouple an upper bound into a knowledge transfer and a spectral regularisation component\footnote{For full details, please see the supplementary material.}:

\begin{small}
\begin{align}
    \mathcal{L}_{distill} &= d(\mathbf{Z}_s, \mathbf{Z}_t\mathbf{P}) \rightarrow \lpnormv{\mathbf{Z}_s - \mathbf{Z}_t\mathbf{P}}_2 && \text{e.g. FitNet loss} \\
    &= \lpnormv{
        \sum_{i \in k}( \bar{\sigma}_i\bar{\mathbf{u}}_i\bar{\mathbf{v}}_i^T - \sigma_i\mathbf{u}_i\mathbf{v}_i^T) + 
        \sum_{i \notin k}\sigma_i\mathbf{u}_i\mathbf{v}_i^T
        }_2 && \text{Low-rank projection} \\
    & \leq \begingroup
    \color{r2d}
        \underbrace{\lpnormv{\sum_{i \in k}
            \bar{\sigma}_i{\bar{\mathbf{u}}_i\bar{\mathbf{v}}_i^T -
            \sigma_i \mathbf{u}_i \mathbf{v}_i^T
        }}_2}_{\text{knowledge transfer}} 
    \endgroup + \begingroup
    \color{s2d}
        \underbrace{\lpnormv{\sum_{i \notin k} \sigma_{i}\mathbf{u}_{i}\mathbf{v}_{i}^T}_2}_{\text{regularisation}},
    \endgroup && \text{Decoupled upper bound}
    \label{eq:decoupled_loss}
\end{align}
\end{small}
where $k$ denotes the set of indices indexing the non-zero singular values of $\bar{\mathbf{Z}}_t$. In practice, any metric $d$ that satisfies the triangle inequality has this decoupled upper bound in the cross-task setting. This result shows that the distillation loss can be decomposed into a knowledge transfer and an implicit spectral regularisation component. It explains how the inverted projection can help to improve performance even when there is little or no knowledge to transfer from the teacher: through a low-rank regularisation on the feature space. We empirically observe this emergent decoupling in figure \ref{fig:eigplots}. Here we see that an inverted projected is more effective at removing irrelevant task information, which is important in the cross-task KD setting.

\vspace{-1em}
\subsection{Teacher-Free Distillation}
The decoupled feature distillation in equation \ref{eq:decoupled_loss} allows us to introduce a novel spectral regularisation loss $\mathcal{L}_{spectral}$.
This loss captures the regularisation effect of the cross-task distillation process without the use of any teacher, therefore we call this method \textit{"teacher-free distillation"} as is similarly done in other works~\cite{Yuan2020RevisitingRegularization}. Its objective is to minimise the least-significant singular vectors of the student model's features while keeping the most-significant.
We define the spectral regularisation loss as follows. Assuming that the singular values/vectors are sorted from most to least significant, the loss can be given as follows:
\begin{equation} 
    \mathcal{L}_{spectral} =
    \lpnormv{\sum_{i = r}^{rank} \sigma_{i}\mathbf{u}_{i}\mathbf{v}_{i}^T}_2,
    \label{eq:teacher-free-spectral-reg}
\end{equation}
where $r$ is a hyperparameter expressing the strength of the regularisation loss and $rank$ is the rank of the student features. More concretely, this hyperparameter defines the number of singular values being preserved. A smaller $r$ will result in more singular values being suppressed, thus leading to a more aggressive regularisation of the feature space. In general, this loss effectively penalises the reconstruction of features by the least significant singular values. It suppresses the features that are overly task-specific, thus forcing the representation into a lower rank space, which leads to better generalisation.
We perform experiments using this loss in section \ref{sec:subsec:experiments-ablation-study-teacher-free-backbones}.

\vspace{-1em}
\section{Experiments and Results}
\label{sec:experiments}

To validate the efficacy of our inverted projector in the cross-task setting, we perform experiments ablating across different distillation methods, task pairs, and architectures. 
We experiment with four target tasks: monocular depth estimation, semantic segmentation, image colourisation, and satellite-to-map translation. For each of these student tasks, teacher tasks are chosen that are either identical, similar, or different to them, thus demonstrating that our method is best-suited for the cross-task case where there are significant differences in the task specific knowledge learned by the teacher and student models.

\noindent\textbf{Randomly-initialised teachers.} 
An interesting question arises when we consider increasingly disparate student and teacher tasks: what happens if a \textit{randomly-initialised} teacher is used? In this case, there is no knowledge shared between the teacher's task and the target task, however the random weights in the teacher may still produce diverse features.
To investigate this question, we also distill from randomly initialised teacher models in our experiments.

\begin{table}[t]
    \centering
    \begin{minipage}[t]{.325\textwidth}
        \centering
        \scriptsize
        \resizebox{\textwidth}{!}{
        \begin{tabular}{lcc}
            \toprule
            \belowrulesepcolor{mygrey} 
            \rowcolor{mygrey} Teacher Task & IoU $\uparrow$ & Pix. Acc. $\uparrow$ \\
            \aboverulesepcolor{mygrey}
            \midrule
            \textit{No teacher} & 34.60 $\pm$ 0.36 & 0.759 $\pm$ 0.001 \\
            \midrule
            Random & \textbf{37.20} & \textbf{0.768} \\
            Classif. & 36.00 & 0.766 \\
            Seg. & 36.50 & 0.767  \\
            \bottomrule
        \end{tabular}
        }
        \vspace{1em}
        \caption*{Semantic segmentation}
        \label{tab:subtab:semseg-results}
    \end{minipage}%
    \hfill
    \begin{minipage}[t]{.325\textwidth}
        \centering
        \scriptsize
        \resizebox{\textwidth}{!}{
        \begin{tabular}{lll}
            \toprule
            \belowrulesepcolor{mygrey} 
            \rowcolor{mygrey} Teacher Task & PSNR $\uparrow$ & FID $\downarrow$ \\
            \aboverulesepcolor{mygrey}
            \midrule
            \textit{No teacher} & 20.48 $\pm$ 0.04 & 65.77 $\pm$ 1.21 \\
            \midrule
            Random & 20.99 & 63.23 \\
            Seg. & 21.10 & 63.44 \\
            Classif & \textbf{21.27} & 65.92 \\
            Depth & 20.84 & \textbf{62.60} \\
            \bottomrule
        \end{tabular}
        }
        \vspace{1em}
        \caption*{Colourisation}
        \label{tab:subtab:pix2pix-results}
    \end{minipage}%
    \hfill
    \begin{minipage}[t]{.325\textwidth}
        \centering
        \scriptsize
        \resizebox{\textwidth}{!}{
        \begin{tabular}{lll}
            \toprule
            \belowrulesepcolor{mygrey} 
            \rowcolor{mygrey} Teacher Task & PSNR $\uparrow$ & FID $\downarrow$ \\
            \aboverulesepcolor{mygrey}
            \midrule
            \textit{No teacher} & 35.29 $\pm$ 0.18 & 67.43 $\pm$ 1.70 \\
            \midrule
            Random & 35.28 & 72.54 \\
            Classif & \textbf{36.29} & \textbf{59.86} \\
            \bottomrule
        \end{tabular}
        }
        \vspace{1em}
        \caption*{Satellite-to-map conversion}
        \label{tab:subtab:sat2map-results-cyclegan}
    \end{minipage}
    \vspace{1em}
    \caption{\textbf{Comparison for different cross-task settings}. We observe that our inverted projector is effective across many different task pairs and even for the same-task settings.}
    \label{tab:subtable}
\end{table}


\subsection{Implementation details}
\label{sec:subsec:implementation-details}
    The general framework used is described in section \ref{sec:method}, and shown in figure \ref{fig:overview-ours-cross-task}. The model architectures used for the student and teacher vary depending on the task pairs. As an example, our depth estimation student is an encoder-decoder architecture with either a MobileNetV2, ResNet50, or EfficientNet-B0 backbone, and the frozen teacher model is a ViT-B/32 \cite{dosovitskiy_image_2021} trained for classification or the SwinV2-B \cite{liu_swin_2022} backbone of AiT \cite{ning_all_2023} pretrained for instance segmentation.
    All architectures used follow an encoder-decoder structure. The student and teacher features, $\mathbf{Z}_s$ and $\mathbf{Z}_t$, are extracted immediately after the encoder of the model in question. All decoders used require features with a spatial (height and width) dimension, therefore if the teacher model's encoder has a final pooling layer (as in the case of the classification teacher), this is removed.

\subsection{Monocular depth estimation}
\label{sec:subsec:experiments-depth}
    Monocular depth estimation is the task of inferring the depth, or distance to the camera, of every point shown in a single image. It is a challenging problem, as there is a many-to-one mapping from 3D scenes to a given 2D depth-image.
    %
    When experimenting with depth estimation as a target task, we make use of teachers trained on tasks that range from similar to dissimilar to the target task of depth estimation, in terms of the overlap in knowledge.
    We use a depth teacher for the same-task distillation, and the instance segmentation and classification tasks for the increasingly dissimilar teacher tasks. Finally, we use a randomly initialised and frozen teacher for the most extreme cross-task setting.
    Experiments are run on the NYUv2 dataset~\cite{silberman_indoor_2012}.
    Our results are shown in table \ref{tab:main_cross_task_distillation}. 
    As expected (see section \ref{sec:method}), the use of our inverted projection produces improvements in performance when using \textit{dissimilar} teacher tasks (random and classification), and gives similar or worse performance when the teacher's task is similar to the target (instance segmentation and depth estimation teachers). 

    We use four different feature distillation methods from the same-task literature to show the utility of our method as a drop-in for use in the cross-task setting: FitNets~\cite{Romero2015FitNets:Nets}, Attention Transfer (AT)~\cite{Zagoruyko2019PayingTransfer}, Probabilistic Knowledge Transfer (PKT)~\cite{Passalis2018LearningTransfer}, and Ensemble (the projector ensemble method of~\cite{Chen2022ImprovedEnsemble}).
    The cross-task improvement using the inverted projection is most pronounced with FitNets and Ensemble, but there is some improvement with AT and PKT. 
    

\begin{figure}[t!]
    \centering
    \subfigure[Input]{\hspace{.19\textwidth}}
    \subfigure[Ground Truth]{\hspace{.19\textwidth}}
    \subfigure[Depth Teacher]{\hspace{.19\textwidth}}
    \subfigure[Insseg Teacher]{\hspace{.19\textwidth}}
    \subfigure[Random Teacher]{\hspace{.19\textwidth}}
    \vspace{-0.5em}
    \subfigure{\includegraphics[width=0.19\textwidth]{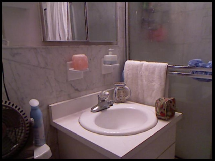}} 
    \subfigure{\includegraphics[width=0.19\textwidth]{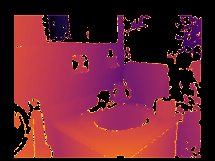}} 
    \subfigure{\includegraphics[width=0.19\textwidth]{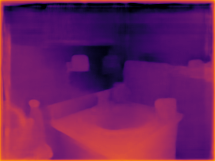}} 
    \subfigure{\includegraphics[width=0.19\textwidth]{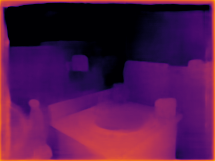}} 
    \subfigure{\includegraphics[width=0.19\textwidth]{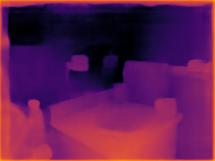}}
    \vspace{-0.5em}
    \subfigure{\includegraphics[width=0.19\textwidth]{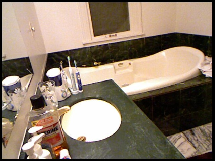}} 
    \subfigure{\includegraphics[width=0.19\textwidth]{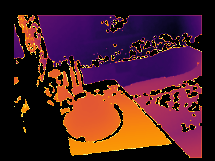}} 
    \subfigure{\includegraphics[width=0.19\textwidth]{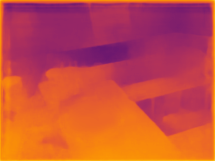}} 
    \subfigure{\includegraphics[width=0.19\textwidth]{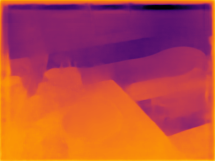}} 
    \subfigure{\includegraphics[width=0.19\textwidth]{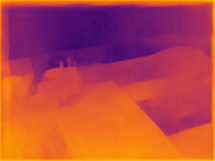}}
    \vspace{1em}
    \caption{\textbf{Qualitative results on NYUv2 (depth) using different teacher tasks:} results from depth estimation, instance segmentation, and randomly-initialised teachers to a MobileNetV2~\cite{Fox2018MobileNetV2:Bottlenecks} student.
    In each case, we use the optimal projection type for the teacher task.}
    \label{fig:qualitative-different-teachers}
\end{figure}

\subsection{Semantic segmentation}
\label{sec:subsec:experiments-seg}
    Semantic segmentation is the task of labelling every pixel in the input image with a class. Our experiments are performed using MSCOCO~\cite{Lin2014MicrosoftContext}, an 80-class segmentation dataset.
    We validate the effectiveness of our inverted projector using segmentation, classification, or randomly initialised teachers. In all experiments, we use a simple L2 loss between the projected teacher features and the student features.
    Results shown in table \ref{tab:subtable}.
    We are able to obtain significant improvement with all the teacher tasks considered, with the best improvements seen with the random teacher. This follows: both classification and semantic segmentation have significant overlap in knowledge, but the random teacher has significant task-irrelevant information that our inverted projection is able to discard.
    This further empirically validates the regularisation components described in equation \ref{eq:decoupled_loss}.

\subsection{Image-to-image translation}
\label{sec:subsec:experiments-im2im}
    We experiment with two image-to-image translation tasks: transforming satellite images into maps, and colourisation of black-and-white images. These two student tasks are particularly important, as they do not have significant knowledge overlap with any of the teacher tasks used: classification, depth estimation, instance segmentation, and the randomly-initialised teacher.
    In contrast, segmentation and classification share a common goal of understanding semantic context of the world, while depth estimation and segmentation have been shown to aid one another (see section \ref{sec:related_work}).
    Results are shown in table \ref{tab:subtable}.
    Given the relative dissimilarity of the student tasks with all teacher tasks, unsurprisingly, our inverted projection performs well in all cases. We are able to use our inverted projection to produce a significant improvement over the baseline with all teacher tasks, including with the randomly initialised teacher. 

\subsection{Teacher-free distillation.}
\label{sec:subsec:experiments-ablation-study-teacher-free-backbones}
    As shown in sections \ref{sec:subsec:experiments-depth}, \ref{sec:subsec:experiments-seg}, and \ref{sec:subsec:experiments-im2im}, we are able to obtain significant performance improvements even when the teacher is randomly initialised and then frozen, thus containing no task-specific knowledge at all.
    This reinforces the conclusion reached in section \ref{sec:subsec:decoupled}: the distillation loss function $\mathcal{L}_{distill}$ may be comprised of a \textit{knowledge transfer} component and a \textit{spectral regularisation component}.
    In the case where there is no knowledge to transfer between the teacher and the student, only a regularising effect can explain the performance improvement over the baseline. 
    To control for this, and to provide further evidence of the loss decoupling described in equation \ref{eq:decoupled_loss}, we perform experiments using the \textit{spectral regularisation loss} (equation \ref{eq:teacher-free-spectral-reg}). Experimenting with different $r$ values in a depth estimation model trained on NYUv2 \cite{silberman_indoor_2012} without a teacher, we find that spectral regularisation significantly enhances performance across all $r$ values, particularly at $r = 2$ (see appendix).
    This supports the decoupling of $\mathcal{L}_{distill}$ into knowledge transfer and regularisation terms (equation \ref{eq:decoupled_loss}) and further validates using a randomly-initialised teacher.
    In table \ref{tab:deit} we consider knowledge distillation on the large-scale ImageNet-1K dataset, and we observe that our simple regularisation loss achieves competitive performance with many state-of-the-art knowledge distillation methods. 
    
    \begin{table}[t]
    \centering
    \footnotesize
    \begin{tabular}{lcc}
        \toprule
        \belowrulesepcolor{mygrey}
        \rowcolor{mygrey} Network & acc@1 & \#params \\
        \aboverulesepcolor{mygrey}
        \midrule
        RegNety 160~\cite{Radosavovic2020DesigningSpaces} & 82.6 & 84M \\
        \midrule
        \multicolumn{3}{l}{\textit{Methods using a pre-trained teacher}} \\
        DeiT-Ti\alambicdeit~\cite{Touvron2021TrainingAttention} & 74.5 & 6M \\
        Co-advise~\cite{Ren2022Co-advise:Distillation} & 74.9 & 6M \\
        DearKD ~\cite{chen2022dearkd} & 74.8 & 6M \\
        USKD~\cite{yang2023knowledge} & 75.0 & 6M \\
        \midrule
        \multicolumn{3}{l}{\textit{Methods \textbf{without} any teacher}} \\
        DeiT-Ti~\cite{Touvron2021TrainingAttention} & 72.2 & 5M\\
        Ours: $\mathcal{L}_{spectral}$$(r=8)$ & \textbf{74.5} & \textbf{5M} \\
        \bottomrule
    \end{tabular}
    \vspace{1em}
    \caption{\textbf{Comparing our novel teacher-free spectral regularisation loss} to other state-of-the-art KD methods on ImageNet-1K~\cite{deng_imagenet:_2009}. 
    Top row is the teacher model used by the KD methods that use a teacher.
    All methods use a DeiT-Ti~\cite{Touvron2021TrainingAttention} student model, with DeiT-Ti\alambicdeit describing the distilled variant using distillation tokens. %
    }
    \label{tab:deit}
\end{table}

\vspace{-1em}
\section{Conclusion}
\label{sec:conclusion}

In this paper we propose the inverted projector as a simple drop-in component for extending many knowledge distillation (KD) methods into cross-task settings, where the teacher's task differs from the student's.
This inverted projector is able to suppress the irrelevant task-specific features from the teacher, which greatly improves the efficacy of cross-task distillation.
We show consistent and substantial improvements across a number of cross-task pairs using our approach. 
Most notably, we achieve up to a 7.47\% improvement for depth estimation by distilling across a significant task-gap. 
Through analysis, we provide a concrete interpretation and explanation for our results, leading to a natural decoupling of the objective into a knowledge transfer and a spectral regularisation component, and we extend this to demonstrate a novel drop-in teacher-free loss that achieves some of the benefits of knowledge distillation without the use of a teacher.
In this work we have highlighted some of the limitations of KD in the cross-task setting, while also providing a step towards broadening its practical applicability in this new domain.

\newpage

\bibliography{references, refs, references-da}



\end{document}